\documentclass[runningheads]{llncs}
\usepackage{graphicx}

\usepackage{tikz}
\usepackage{comment}
\usepackage{amsmath,amssymb} 
\usepackage{color}

\definecolor{Gray}{gray}{0.95}
\usepackage{colortbl}
\newcolumntype{a}{>{\columncolor{Gray}}c}

\newcommand{\seasons}[1]{\textcolor{black}{#1}}
\newcommand{\ljx}[1]{\textcolor{black}{#1}}
\usepackage[accsupp]{axessibility}  

\usepackage[linesnumbered,lined,vlined,ruled,commentsnumbered]{algorithm2e}
\usepackage{graphicx}
\usepackage{amsmath}
\usepackage{amssymb}
\usepackage{booktabs}
\usepackage{times}
\usepackage{helvet}
\usepackage{courier}
\usepackage{epsfig}
\usepackage{mathrsfs}
\usepackage{multirow}
\usepackage{color}
\usepackage{xcolor}
\usepackage{colortbl}
\usepackage{tabularx}
\usepackage{gensymb}
\usepackage{arydshln}
\usepackage{bbding} 
\usepackage{url}
\usepackage{wrapfig}

\begin{document}
\pagestyle{headings}
\mainmatter
\def\ECCVSubNumber{2494}  

\title{tSF: Transformer-based Semantic Filter for Few-Shot Learning} 

\titlerunning{tSF: Transformer-based Semantic Filter for Few-Shot Learning}
%
\author{Jinxiang Lai \and
Siqian Yang \and
Wenlong Liu \and
Yi Zeng \and Zhongyi Huang \and
Wenlong Wu \and Jun Liu \and
Bin-Bin Gao \and Chengjie Wang\thanks{Corresponding Author}}
%
\authorrunning{Jinxiang Lai et al.}
%
\institute{Youtu Lab, Tencent \\
\email{\{jinxianglai, seasonsyang, sylviazeng, ezrealwu, jasoncjwang\}@tencent.com, \{nicehuster, huangzhny, junsenselee, csgaobb\}@gmail.com}}
\maketitle

\begin{abstract}
Few-Shot Learning (FSL) alleviates \seasons{the data shortage challenge} via \seasons{embedding} \seasons{discriminative} target-aware features \seasons{among} plenty seen (base) and few unseen (novel) labeled samples.
Most feature embedding \seasons{modules} in recent FSL methods are specially designed for \seasons{corresponding} learning tasks (e.g., classification, segmentation, and object detection), which limits the \seasons{\emph{utility}} of \seasons{embedding} features.
To this end, we propose a light and universal module named transformer-based Semantic Filter (tSF), \seasons{which can be applied for different FSL tasks}.
\seasons{The proposed tSF redesigns the inputs of a transformer-based structure by a semantic filter, which not only embeds the knowledge from whole base set to novel set but also filters semantic features for target category.}
Furthermore, the parameters of tSF is equal to half of a standard transformer block (less than $1M$).
In the experiments, our tSF is able to boost the performances in different classic few-shot learning tasks (about $2\%$ improvement), especially outperforms the state-of-the-arts on multiple benchmark datasets in few-shot classification task.
\end{abstract}

\section{Introduction}
\label{sec:intro}
Few-Shot Learning (FSL) aims to recognize unseen \seasons{objects} with plenty known data (base) and few labeled unknown samples (novel).
Due to the shortage of novel data, \seasons{FSL tasks} suffer from weak representation problem.
\seasons{Hence, researchers \cite{hou2019cross,yan2019simpleshot,tian2020rethinking,rizve2021exploring,zhengyu2021pareto,zhiqiang2021partial,liu2021learning,xu2021learning} manage to design a embedding network to make extracted features robust and fine-grained enough in unseen instances recognization.}
\seasons{To deal with} different FSL tasks (e.g., classification, segmentation, and object detection), researchers propose different feature embedding \seasons{modules}, e.g., FEAT \cite{ye2020feat}, CTX \cite{doersch2020crosstransformers} for classification, HSNet \cite{min2021hypercorrelation} for segmentation, and FSCE \cite{sun2021fsce} for detection, respectively.

\begin{figure}[!t]
\centering
\includegraphics[width=1.0\linewidth]{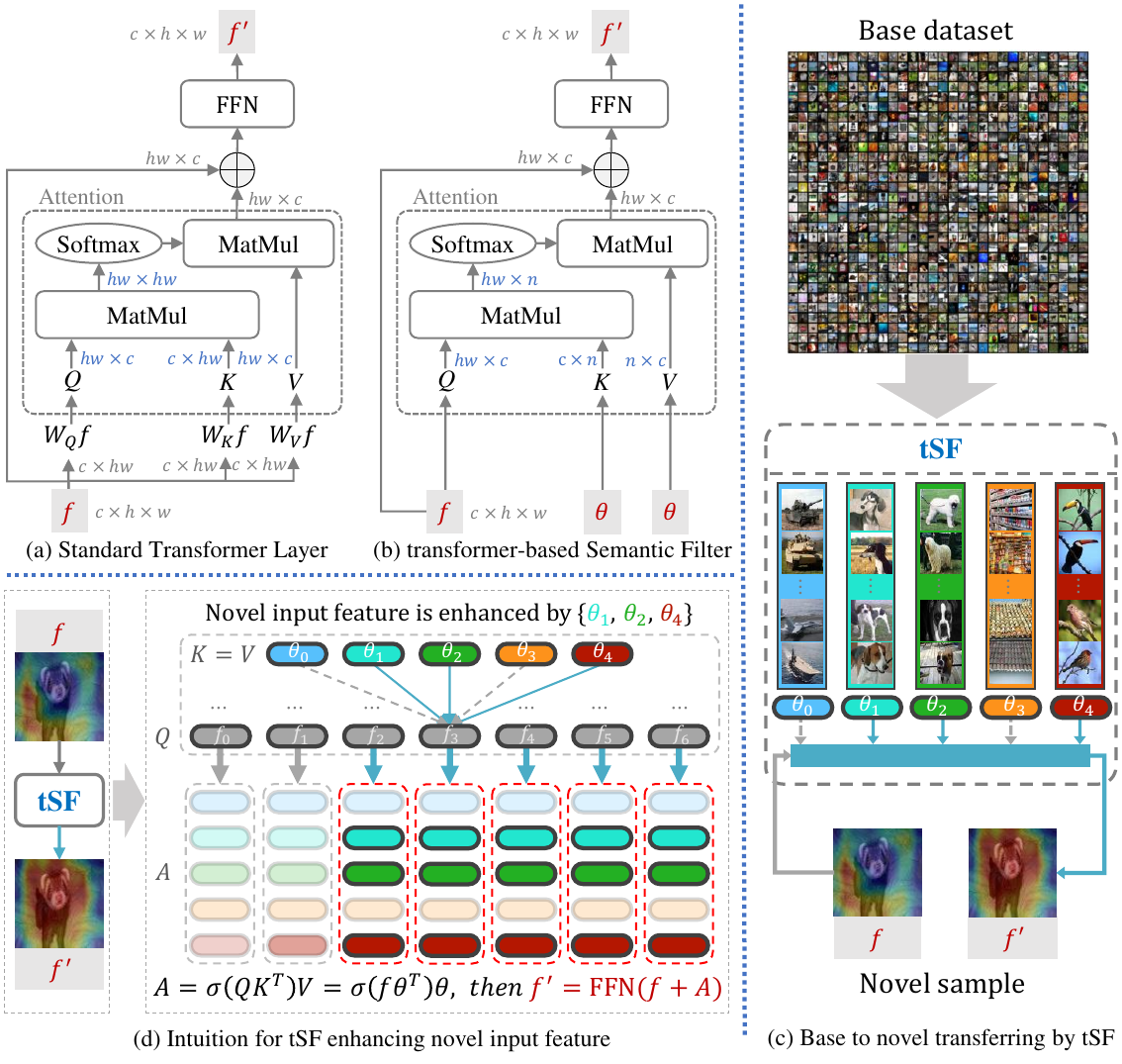}
\vspace{-4mm}
\caption{
\ljx{(a) Standard Transformer layer. (b) transformer-based Semantic Filter (tSF), where ${\theta}$ is learnable semantic filter.
(c) Base to novel transferring by tSF. After training, the semantic info of base dataset are embedded into ${\theta}$, e.g. there are $n=5$ semantic groups and ${\theta}_1$ represents dog-like group. Then, given a novel input sample, tSF enhances its regions which are semantic similar to ${\theta}$.
(d) Intuition for tSF enhancing novel input feature.}}
\label{fig:tSF}
\vspace{-4mm}
\end{figure}

Nevertheless, these methods are limited by the purposes of different tasks.
In the classification task, methods put emphasis on locating the representive prototype for each class.
Methods in the detection task \seasons{aim} to distinguish the similar objects and correct the bounding box, while in \seasons{the} segmentation task, methods \seasons{manage} to generate a precise mask.
In consequence, classification task requires more robust features, while detection task and segmentation task need more fine-grained features.
To satisfy the demands of different tasks \seasons{concurrently}, \emph{a target-aware and image-aware feature embedding method is inevitable}.

Throughout the recent investigations, transformer-base structure \cite{vaswani2017attention} brings an important significance at computer vision field, which works on almost all common tasks due to its sensibility on both big dataset (macro) and a single image (micro).
Specifically, transformer is able to store the information of whole dataset  modeling spatio-temporal correlations among instances. The property exactly satisfies the purpose of the feature embedding operation.
Besides, observing from Transformer \cite{vaswani2017attention}, Feat \cite{ye2020few}, SuperGlue \cite{sarlin2020superglue}, CTX \cite{doersch2020crosstransformers} and DETR \cite{carion2020end}, we notice that the transformer layer could perform different learning behaviors with different input forms of ${\{Q,K,V\}}$.
In common, a traditional transformer structure needs big training dataset to achieve high performances.
However, in few-shot learning \seasons{field}, it may fall into overfitting problem due to the shortage of data without a carefully designed framework.

To this end, we propose a light and general feature embedding module, named transformer based Semantic Filter (tSF), as illustrated in Fig.~\ref{fig:tSF}(b).
We redesign the inputs ${Q, K, V}$ of traditional transformer as ${f, \theta, \theta}$, where $f$ is the extracted feature and $\theta$ is a learnable weight, named semantic filter.
The tSF uses the correlation matrix between $(f,\theta)$ to re-weight $\theta$.
The average of the re-weighted $\theta$ is involved the dataset-attention response, which enhances the feature $f$ from the views of both global dataset and local image.
In this way, tSF can be trained without big data, while keeping the macro and micro information at the same time.
\ljx{Intuitively, Fig.~\ref{fig:tSF}(c) and Fig.~\ref{fig:tSF}(d) show that the proposed tSF is able to enhance the foreground regions of input novel feature via the embedded semantic info of base dataset.}
Besides, to further show the efficient of tSF, we insert the tSF into a strong few-shot classification framework, named PatchProto.
The experimental results show that tSF helps PatchProto achieve SOTA performance with about $2\%$ improvement.
In addition, the parameter size of tSF is less than $1 M$, half of that in traditional transformer.
Moreover, to prove that tSF can suit different FSL frameworks, we conduct massive experiments on different few-shot learning tasks, and the results show that tSF can make $2\%-3\%$ improvements on classification, \seasons{detection and segmentation} tasks.

In summary, our contributions are listed as follows:

$\bullet$ An effective and light module named transformer-based Semantic Filter (tSF) is proposed, which is helpful to learn a generalized-well embedding for novel targets (unseen in model training phase). The tSF leverages dataset-attention mechanism to realize information interaction between single input instance and whole dataset.

$\bullet$ A strong baseline framework called PatchProto is introduced for few-shot classification. Experimental results show our PatchProto+tSF approach outperforms the state-of-the-arts on multiple benchmark datasets such as miniImageNet and tieredImageNet.

$\bullet$ As a universal module, tSF is applied in different few-shot learning tasks, including classification, semantic segmentation and object detection. The massive experiments demonstrate that tSF is able to boost their performances.

\section{Related Work}
\label{sec:FSC_related_work}
\noindent\textbf{Few-Shot Classification} \ \
FSL algorithms first pre-train a base classifier with abundant samples \seasons{(seen images)}, then learn to recognize novel classes (unseen images) with a few labeled samples.
According to recent investigations, there are four representative directions: optimization-based, parameter-generating based, metric-learning based, and embedding-based methods.
\emph{Optimization-based methods} are able to perform rapid adaption with a few training samples for new classes by learning a good optimizer \cite{ravi2016optimization,marcin2018learn} or learning a well-initialized model \cite{finn2017model,nichol2018first,rusu2019meta}.
\emph{Parameter-generating methods} \cite{bertinetto2016learning,munkhdalai2017meta,qi2018memory,munkhdalai2018rapid,gidaris2019generating} focus on learning a parameter generating network.
\emph{Metric-learning based methods} learn to compare to tackle the few-shot classification problem.
The main idea is classifying a new input image by computing the similarity compared with labeled instances \cite{gregory2015siamese,vinyals2016matching,hou2019cross,xu2021learning,snell2017prototypical,sung2018learning,wu2019parn,zhang2020deepemd}.
These methods design carefully to embedding network to match their corresponding distance metrics.
\emph{Embedding-based methods} \cite{yan2019simpleshot,tian2020rethinking,rizve2021exploring,zhengyu2021pareto,zhiqiang2021partial,liu2021learning} firstly focus on learning a generalize-well embedding with supervised or self-supervised learning tasks, and then freeze this embedding and further train a linear classifier or design a metric classifier on novel classes.

\noindent\textbf{Auxiliary Task for Few-shot Classification} \ \
Some recent works gain a performance improvement by training few-shot models with supervised and self-supervised auxiliary tasks.
The supervised task for FSL simply performs global classification on the base dataset as in CAN \cite{hou2019cross}.
The effectiveness of self-supervised learning for FSL has been demonstrated, such as contrastive learning in either unsupervised pre-training \cite{carlos2020self} or episodic training \cite{doersch2020crosstransformers,liu2021learning}, and auxiliary rotation prediction task \cite{gidaris2019boosting,su2020does,rizve2021exploring}.

\noindent\textbf{Few-Shot Semantic Segmentation} \ \
Early few-shot semantic segmentation methods apply a dual-branch architecture \cite{shaban2017one,dong2018few,rakelly2018conditional}, one segmenting query-images with the prototypes learned by the other branch. In recently, the dual-branch architecture is unified into a single-branch, using the same embedding for support and query images \cite{zhang2020sg,siam2019amp,wang2019panet,rpmm,ppnet}. These methods aim to leverage better guidance for the segmentation of query-images \cite{zhang2020sg,nguyen2019feature,wangfew,zhang2019pyramid}, via learning better class-specific representations \cite{wang2019panet,liu2020crnet,ppnet,rpmm,siam2019amp} or iteratively refining \cite{zhang2019canet}.

\noindent\textbf{Few-Shot Object Detection} \ \
Existing few-shot object detection approaches can be divided into two paradigms: meta-learning based \cite{kang2019few,xiao2020few,fan2020few,hu2021dense} and transfer learning based \cite{wang2020frustratingly,wu2020multi,sun2021fsce,fan2021generalized,qiao2021defrcn}.
The majority of meta-learning approaches adopt \emph{feature reweighting} or its variants to aggregate query and support features, which predict detections conditioned on support sets. Differently, the transfer learning based approaches firstly train the detectors on base set, then fine-tune the task-head layer on novel set, which achieve competitive results comparing to meta-learning approaches.

\noindent\textbf{Transformer} \ \
Transformer is an attention-based network architecture that is widely applied in natural language processing area \cite{vaswani2017attention,devlin2018bert}. Due to its power in learning representation, it has been introduced in many computer vision tasks, such as image classification \cite{dosovitskiy2020image,wang2021pyramid,touvron2021training}, detection \cite{zhang2020feature,carion2020end,zhu2020deformable}, segmentation \cite{zheng2021rethinking,liang2020polytransform,xie2021segformer}, image matching \cite{sun2021loftr,sarlin2020superglue} and few-shot learning \cite{doersch2020crosstransformers,yang2020context,lu2021simpler}.

To best of our knowledge, there is no general feature embedding method, which can be applied on multiple few-shot learning tasks.

\section{Transformer-based Semantic Filter (tSF)}
\subsection{Related Transformer}
\label{sec:tSF_related_work}
\noindent\textbf{Transformer-based Self-Attention} \ \
The architecture of a standard Transformer \cite{vaswani2017attention} layer is presented in Fig.~\ref{fig:tSF}(a), which consists of Attention and Feed-Forward Network (FFN).
Given input feature ${f \in \mathbb{R}^{c\times h\times w}}$, the Transformer layer outputs ${f^{'} \in \mathbb{R}^{c\times h\times w}}$.
The key operation \emph{Attention} is expressed as:
\begin{equation}
\vspace{-0.5mm}
Attention(Q,K,V)=\sigma(QK^T)V,
\label{equ:attention}
\vspace{-0.5mm}
\end{equation}
where, ${\{Q,K,V\}}$ are known as query, key and value respectively,  ${\sigma}$ is softmax function. The forms of ${\{Q,K,V\}}$ in \emph{transformer-based self-attention} are:
\begin{equation}
\vspace{-0.5mm}
Q=W_{Q}f,\quad K=W_{K}f,\quad V=W_{V}f,
\label{equ:QKV_self}
\vspace{-0.5mm}
\end{equation}
where, ${\{Q,K,V\} \in \mathbb{R}^{hw \times c }}$ and some feature-reshaping operations are omitted for simplicity, ${W_{Q},W_{K},W_{V}}$ are learnable weights (e.g. convolution layers).
For few-shot classification, based on self-attention mechanism, Feat \cite{ye2020few} used the standard transformer layer as a set-to-set function to perform embedding adaptation, formally:
\begin{equation}
\vspace{-0.5mm}
Q=W_{Q}f_{set},\quad K=W_{K}f_{set},\quad V=W_{V}f_{set},
\label{equ:QKV_feat}
\vspace{-0.5mm}
\end{equation}
where ${f_{set}}$ is a set of features of all the instances in the support set.
Both the standard transformer \cite{vaswani2017attention} and Feat \cite{ye2020few} are based on self-attention mechanism which learns to model the relationship between the feature-nodes insight the input features.
Differently, as illustrated in Fig.~\ref{fig:tSF}(b) and Eq.~\ref{equ:QKV_tSF}, our Transformer-based Semantic Filter (tSF) applies \emph{Transformer-based Dataset-Attention}, which makes information interaction between single input sample and whole dataset.

\noindent\textbf{Transformer-based Cross-Attention} \ \
The standard Transformer performs self-attention behavior, and SuperGlue \cite{sarlin2020superglue} found that Transformer can be used to make cross-attention between pair-features. Given input pair-features (${f_{ref},f}$), it can obtain a cross-correlation matrix between (${f_{ref},f}$) which is used to re-weight ${f}$ and achieves the cross-attention implementation.
Specifically, ${\{Q,K,V\}}$ forms in \emph{transformer-based cross-attention} are:
\begin{equation}
\vspace{-0.5mm}
Q=W_{Q}f_{ref},\quad K=W_{K}f,\quad V=W_{V}f.
\label{equ:QKV_cross}
\vspace{-0.5mm}
\end{equation}
For few-shot classification, based on cross-attention mechanism, CTX \cite{doersch2020crosstransformers} used the transformer layer to generate query-aligned prototype for support class.

\noindent\textbf{Transformer-based Decoder} \ \
The forms of ${\{Q,K,V\}}$ in transformer-based decoder of DETR \cite{carion2020end} are:
\begin{equation}
\vspace{-0.5mm}
Q=\varphi,\quad K=V=f,
\label{equ:QKV_DETR}
\vspace{-0.5mm}
\end{equation}
where ${\varphi \in \mathbb{R}^{u\times c}}$ is learnable weights which is called as object queries. Given input feature ${f \in \mathbb{R}^{c\times h\times w}}$, the DETR \cite{carion2020end} decoder outputs ${\varphi^{'} \in \mathbb{R}^{u\times c}}$ which is used to locate the objects. The dimension ${u}$ of object queries ${\varphi}$ represents the maximum number of objects insight a image.

\subsection{tSF Methodology}
\begin{figure}[!t]
\centering
\includegraphics[width=0.99\linewidth]{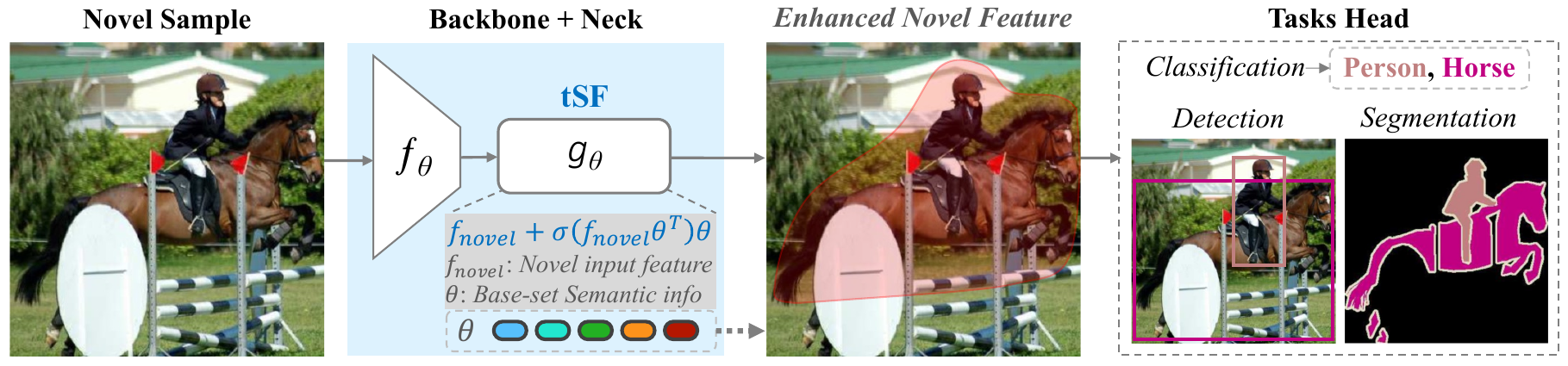}
\vspace{-2mm}
\caption{The tSF for few-shot learning tasks such as classification, semantic segmentation and object detection.}
\label{fig:tSF_app}
\vspace{-4mm}
\end{figure}

In few-shot learning task, obtaining a generalized-well embedding for novel categories is one of the key problem.
To this end, we plan to model the whole dataset information and then transfer the knowledge from base set to novel set.
Benefiting from the property of modeling spatio-temporal correlations among instances, transformer is able to learn the whole dataset information.
Besides, observing from Transformer \cite{vaswani2017attention}, Feat \cite{ye2020few}, SuperGlue \cite{sarlin2020superglue}, CTX \cite{doersch2020crosstransformers} and DETR \cite{carion2020end}, we notice that the transformer layer performs different learning behavior with different input forms of ${\{Q,K,V\}}$.
Therefore, in order to transfer the knowledge from whole base set to novel set, we propose a transformer-based Semantic Filter (tSF) as illustrated in Fig.~\ref{fig:tSF}(b), where the forms of ${\{Q,K,V\}}$ are designed as:
\begin{equation}
\vspace{-0.5mm}
Q=f,\quad K=V=\theta,
\label{equ:QKV_tSF}
\vspace{-0.5mm}
\end{equation}
where, ${Q \in \mathbb{R}^{hw \times c }}$ after feature-reshaping, and ${\theta \in \mathbb{R}^{n\times c}}$ is learnable weights which is called as semantic filter.
\ljx{
Formally, tSF is expressed as:
\begin{equation}
\vspace{-0.5mm}
{f^{'} = tSF(f) = FFN\left(f + \sigma\left(f {\theta}^T\right){\theta}\right)}.
\label{equ:full_tSF}
\vspace{-0.5mm}
\end{equation}
The input and output features of tSF are ${\{f,f^{'}\} \in \mathbb{R}^{c\times h\times w}}$ respectively, which is consistent with the standard Transformer. The tSF can be utilized as a feature-to-feature function, e.g. stacking tSF as a neck after the backbone architecture for few-shot learning as illustrated in Fig.~\ref{fig:tSF_app} and Fig.~\ref{fig:PatchProto}(a).
}

\ljx{Formally, let's define $C_{base}$ and $C_{novel}$ as categories of base and novel respectively.
Although ${C_{base} \cap C_{novel} = \emptyset}$, we assume $C_{base}$ consists of sub-sets $C_{base}^{sim}$ and $C_{base}^{diff}$ which are semantically similar and different from $C_{novel}$ respectively.
Then, tSF transfers knowledge from base to novel:
\begin{equation}
\vspace{-0.5mm}
\begin{aligned}
&Base \ Training: \ {f^{'}_{base} = FFN\left(f_{base}+\sigma\left(f_{base} {\theta}^T\right){\theta}\right)}, \\
&Novel \ Testing: \ {f^{'}_{novel} = FFN\left(f_{novel}+\sigma\left(f_{novel} {\theta}^T\right){\theta}\right)}.
\end{aligned}
\label{equ:b2n_tSF}
\vspace{-0.5mm}
\end{equation}
After training on base, semantic info of $C_{base}$ are embedded into $\theta$ of which dimension $n$ are interpreted as projected semantic groups, i.e. $\theta$ also consists of sub-sets $\theta^{sim}$ and $\theta^{diff}$ which are similar and different from $C_{novel}$ respectively.
Given a novel image, tSF enhances its regions which are similar to $\theta^{sim}$ while $\theta^{diff}$ doesn't.
For example, if 'dog' in base, tSF enhances novel 'wolf' due to their high similarity relation.
In Fig.~\ref{fig:tSF}(d), $\theta^{sim}$ is $\{\theta_1, \theta_2, \theta_4\}$, $\theta^{diff}$ is $\{\theta_0, \theta_3\}$.
}

Intuitively, as illustrated in Fig.~\ref{fig:tSF}(c), with the model training on base set, the semantic filter ${\theta}$ learns the whole dataset information.
In the model testing on novel set, according to Eq.~\ref{equ:full_tSF} and Fig.~\ref{fig:tSF}(d), the tSF uses the correlation matrix between ${(f,{\theta})}$ to re-weight ${\theta}$. The average of the re-weighted ${\theta}$ is the dataset-attention response ${A}$, which is semantically similar to ${f}$ and can be used to enhance the target object as ${f^{'} = FFN(f+A)}$.
${f}$ and ${\theta}$ contains the info of one sample and whole dataset respectively, and the tSF can collect the target information (i.e. A which is semantically similar to ${f}$) from ${\theta}$ to enhance ${f}$.
Therefore the foreground region of the input novel sample feature can be enhanced by the dataset-attention response.

\subsection{Discussions}
\noindent\textbf{Visualizations} \ \ Under the few-shot classification framework of PatchProto+tSF introduced in Sec.~\ref{subsec:patchproto_tsf}, we give the visualizations of feature response map and t-SNE for tSF as shown in Fig.~\ref{fig:PatchProto}(c) and Fig.~\ref{fig:PatchProto}(b) respectively.
In detail, ${f^{'}=g_{\theta}(f)}$ and ${g_{\theta}}$ is the proposed tSF, the correlation matrix ${R=\sigma(f_{novel} {\theta}^T) \in \mathbb{R}^{hw\times n}}$ and ${R\_{\theta_i}\in \mathbb{R}^{hw}}$ represents the correlation vector between ${f_{novel}}$ and ${\theta_i \in \mathbb{R}^{c}}$ (${i^{th}}$ position of ${\theta}\in \mathbb{R}^{n\times c}$).
In Fig.~\ref{fig:PatchProto}(c), comparing to ${f}$, ${f^{'}}$ obtains more accurate and complete response map focusing on the foreground region, which indicates that tSF is able to transfer semantic knowledge from base set to novel set.
The visualizations of the correlation vector ${R\_{\theta_i}}$ show that ${\theta_i}$ learns semantic information from base set,
\ljx{specifically, these targets foreground are enhanced mainly contributed from ${\{\theta_1,\theta_2,\theta_4\}}$.}
In addition, the t-SNE visualizations in Fig.~\ref{fig:PatchProto}(b) show that ${f^{'}}$ obtains more clear category boundaries than ${f}$, which demonstrates that tSF is able to produce more discriminative embedding features for novel categories.

\noindent\textbf{Properties} \ \ The properties of tSF are as follows:
(i) Generalization ability: Based on dataset-attention mechanism, the tSF models the whole dataset information and then transfer the knowledge from base set to novel set. The tSF makes information interaction between input sample and whole dataset, while self-attention based transformer interacts info insight input sample itself which leads to overfitting problem due to insufficient information interaction.
(ii) Representation ability: The low dimension semantic filter ${\theta}\in \mathbb{R}^{n\times c}$ learns high-level semantic information from whole dataset.
(iii) Efficiency: The computational complexity of tSF is less than self-attention based transformer. The complexity of transformer in calculating \emph{Attention} by Eq.~\ref{equ:attention} is ${O(transformer)=(h\times w)^{3} \times c}$, while our tSF is only ${O(tSF)=(h\times w)^{2} \times n \times c}$, i.e. ${\frac{O(transformer)}{O(tSF)}=\frac{h\times w}{n}\gg{1}}$.

\noindent\textbf{Comparisons} \ \
\ljx{Comparing our tSF with Transformer and DETR decoder, in model testing on novel set, the input novel feature ${f_{novel}}$ is enhanced by different information.
Our tSF enhances ${f_{novel}}$ by base dataset info (i.e. the semantic filter ${\theta}$ learned on base set) as defined in Eq.~\ref{equ:b2n_tSF}, which fulfils base to novel transferring.
Differently, the standard Transformer enhances ${f_{novel}}$ by itself instance info:
\begin{equation}
\vspace{-0.5mm}
Transformer:\quad {f^{'}_{novel} = FFN\left(f_{novel}+\sigma\left(f_{novel} {f_{novel}}^T\right){f_{novel}}\right)}.
\label{equ:fnovel_transformer}
\vspace{-0.5mm}
\end{equation}
Besides, the DETR decoder also enhances ${f_{novel}}$ by itself instance info:
\begin{equation}
\vspace{-0.5mm}
DETR:\quad {f^{'}_{novel} = FFN\left(f_{novel}+\sigma\left(\theta {f_{novel}}^T\right){f_{novel}}\right)}.
\label{equ:fnovel_transformer}
\vspace{-0.5mm}
\end{equation}
As illustrated in Tab. \ref{table:ablation_1}, the experimental comparisons show that our tSF obtains obvious performance gains, while Transformer and DETR show performance degradation due to overfitting problem.
}

\section{tSF for Few-Shot Classification}

\subsection{Problem Definition}
\label{sec:FSC_ProblemDef}
${N}$-way ${M}$-shot task to learn a classifier for $N$ unseen classes with $M$ labeled samples.
Formally, we have three mutually disjoint datasets: a base set ${X_{base}}$ for training, a validation set ${X_{val}}$, and a novel set ${X_{novel}}$ for testing.  Following \cite{vinyals2016matching,sung2018learning,hou2019cross,xu2021learning}, the episodic training strategy is adopted to mimic the few-shot learning setting, which has shown that it can effectively train a meta-learner (i.e., a few-shot classification model).
Each episode contains $N$ classes with $M$ samples per class as the support set $\mathcal{S}=\{\left(x^s_i, y^s_i\right)\}_{i=1}^{m_s}$ ($m_s=N\times M$), and a fraction of the rest samples as the query set $\mathcal{Q}=\{\left(x^q_i, y^q_i\right)\}_{i=1}^{m_q}$. And the support subset of the $k^{th}$ class is denoted as $\mathcal{S}^k$.

\subsection{PatchProto Framework with tSF}
\label{subsec:patchproto_tsf}

\begin{figure}[!t]
\centering
\includegraphics[width=0.99\linewidth]{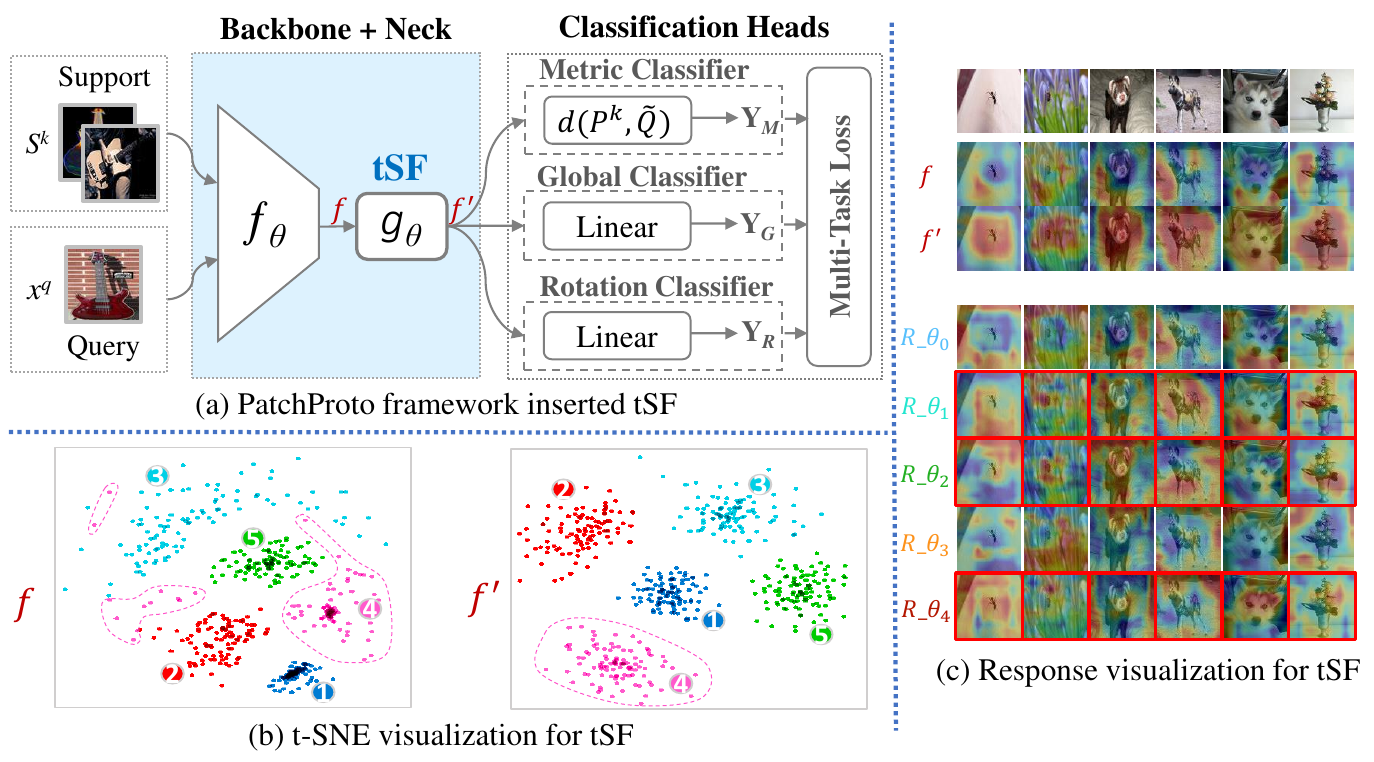}
\vspace{-2mm}
\caption{
\ljx{(a) The PatchProto framework inserted tSF for few-shot classification.
(b) The t-SNE visualization comparison for PatchProto+tSF, where ${f^{'}=g_{\theta}(f)}$ and ${g_{\theta}}$ is the proposed tSF.
(c) The visualizations of response map with the input of novel sample, where ${R\_{\theta_i}}$ is the correlation vector between ${(f,\theta_i)}$, and the dimension ${n}$ of ${\theta \in \mathbb{R}^{n\times c}}$ is set to 5.}}
\label{fig:PatchProto}
\vspace{-4mm}
\end{figure}

As illustrated in Fig.~\ref{fig:PatchProto}, the proposed PatchProto network consists of five components: feature extractor backbone ${f_\theta}$, transformer-based Semantic Filter (tSF) ${{g}_\theta}$, Metric classifier ${f_M}$ for few-shot classification, and Global ${f_G}$ and Rotation ${f_R}$ classifiers for auxiliary tasks which are only used to assist model training.

The input image $x^q$ in the query set $\mathcal{Q}=\{\left(x^q_i, y^q_i\right)\}_{i=1}^{m_q}$ is rotated with [0\degree, 90\degree, 180\degree, 270\degree] and outputs a rotated $\mathcal{\tilde{Q}}=\{\left(\tilde{x}^q_i, \tilde{y}^q_i\right)\}_{i=1}^{{m_q}\times4}$. Each support subset $\mathcal{S}^k$ and a rotated query image $\tilde{x}^q$ are fed through the feature extractor backbone ${f_\theta}$ and tSF ${g_\theta}$, and produces the class feature $P^k=\frac{1}{|\mathcal{S}^k|} \sum_{x^s_i\in \mathcal{S}^k} {g_\theta}({f_\theta}(x^s_i))$ and query feature $\tilde{Q}={g_\theta}({f_\theta}(\tilde{x}^q))\in \mathbb{R}^{c\times h\times w}$, respectively. Then the Metric classifier ${f_M}$ makes classification via measuring the similarity between each pair-features ($P^k$, $\tilde{Q}$). Finally, PatchProto network is trained by optimizing the multi-task loss contributing from meta loss and auxiliary loss.
In inductive inference phase, with the embedding learned in train set, the Metric classifier predicts the query $x^q$ as ${Y_M}$ based on cosine similarity measurement.

\subsubsection{Objective functions}

\noindent\emph{Meta loss:} As a metric-based meta learner, the Metric classifier predicts the query into $N$ support categories by measuring cosine similarity. Following \cite{hou2019cross}, we adopt the patch-wise classification mechanism to generate precise embeddings. Specifically, each local feature $\tilde{Q}_m$ at $m^{th}$ spatial position of $\tilde{Q}$, is predicted into $N$ categories.
Formally, the probability of recognizing $\tilde{Q}_m$ as $k^{th}$ category is:
\begin{equation}
\vspace{-0.5mm}
\hat{Y}(y=k|\tilde{Q}_m)=\sigma\left({d\left(\tilde{Q}_m, \textit{GAP}\left({P}^k\right)\right)} \right),
\label{equ:pred}
\vspace{-0.5mm}
\end{equation}
where \textit{GAP} denotes global average pooling, ${d}$ is cosine distance. Then the metric classification loss with the few-shot label $\tilde{y}^q$ is:
\begin{equation}
\vspace{-0.5mm}
\mathcal{L}_M = -\sum_{i=1}^{m_q} \sum_{m=1}^{h \times w}\log \hat{Y}(y=\tilde{y}^q_i|(\tilde{Q}_m)_i).
\label{equ:LM}
\vspace{-0.5mm}
\end{equation}

\noindent\emph{Auxiliary loss:} The Global classifier predicts the query into all $C$ categories of train set, thus its loss is:
\begin{equation}
\vspace{-0.5mm}
\begin{aligned}
\mathcal{L}_G&=PCE(\tilde{Q},C^q)=-\sum_{i=1}^{m_q} \sum_{n=1}^{h \times w} {C^q_i} \log \left(\sigma(W_l(\tilde{Q}_m)_i)\right).
\end{aligned}
\label{equ:LG}
\vspace{-0.5mm}
\end{equation}
where, $W_l$ is a linear layer, ${C^q_i}$ is the global category of $\tilde{x}^q_i$ with all $C$ categories, and ${PCE}$ denotes the patch-wise cross-entropy function. Similarly, the loss of Rotation classifier is derived by ${\mathcal{L}_R=PCE(\tilde{Q},B^q)}$, where ${B^q_i}$ is the rotation category of $\tilde{x}^q_i$ with four categories.

\noindent\emph{Multi-task loss:} Therefore, inspired by \cite{lai2022rethinking}, the overall classification loss is:
\begin{equation}
\vspace{-0.5mm}
\begin{aligned}
\mathcal{L} = \frac{1}{2}{\mathcal{L}_M} + \sum_{j=G,R}\left({\left({\lambda}+{w_j}\right)}{\mathcal{L}_j}+{log{\frac{1}{{({\lambda}+{w_j})}}}}\right),
\end{aligned}
\label{equ:Loss}
\vspace{-0.5mm}
\end{equation}
where, ${w} = \frac{1}{2{\alpha^2}}$, ${\alpha}$ is learnable variable, ${\lambda}$ is a hyper-parameter to balance the effects of the losses of few-shot task and auxiliary tasks. The influence of ${\lambda}$ is studied in Tab.~\ref{table:ablation_2}.

\section{tSF for Few-Shot Segmentation and Detection}
As shown in Fig.~\ref{fig:tSF_app}, the proposed tSF is stacked after the backbone architecture (i.e. Backbone + tSF) for few-shot learning tasks, such as classification, segmentation and detection. To verify the effectiveness and the university of our tSF module, we insert the tSF into the current state-of-the-art few-shot segmentation and detection methods. And the details are introduced in the next.

\noindent\textbf{RePRI+tSF for Segmentation} \ \
RePRI (Region Proportion Regularized Inference) \cite{Malik2021repri} approach leverages the statistics of unlabeled pixels for the input image. It optimizes three complementary loss terms, including cross-entropy on labeled support pixels, Shannon entropy on unlabeled query pixels and a global KL-divergence regularizer on predicted foreground.
RePRI achieves state-of-the-art on few-shot segmentation benchmark PASCAL-5$^i$ built from PASCAL VOC \cite{everingham2010pascal}.
Based on the RePRI framwork, we simply stack our tSF behind its backbone which obtains the RePRI+tSF approach.

\noindent\textbf{DeFRCN+tSF for Detection} \ \
DeFRCN (Decoupled Faster R-CNN) \cite{qiao2021defrcn} is a simple yet effective fine-tuned approach for few-shot object detection, which proposes Gradient Decoupled Layer and Prototypical Calibration Block to alleviate the contradictions of Faster R-CNN in FSL scenario. Due to its simplicity and effective, DeFRCN achieves state-of-the-art on PASCAL VOC \cite{everingham2010pascal} and COCO \cite{lin2014microsoft}. To verify the effectiveness of tSF module in few shot object detection task, we use DeFRCN as baseline, and insert the tSF into the ResNet-101 backbone to obtain the DeFRCN+tSF approach, specifically, tSF module is followed in the 5th residual block. For the hyperparameter settings, we use the default parameter as same with DeFRCN.

\section{Experiments}
To prove the effectiveness and universality of the proposed tSF, massive experiments are conducted on differnet few-shot learning tasks, including classification, semantic segmentation and object detection. Overall, the results show that tSF can make $2\%-3\%$ improvements on these tasks.

\subsection{Few-Shot Classification}
\noindent\textbf{Datasets} \ \
\emph{mini}ImageNet dataset is a subset of ImageNet \cite{krizhevsky2012imagenet}, which consists of 100 classes. We split the 100 classes following the setting in
\cite{sung2018learning,hou2019cross,xu2021learning}, i.e. 64, 16 and 20 classes for training, validation and test respectively. \emph{tiered}ImageNet dataset \cite{ren2018meta} is also a subcollection of ImageNet \cite{krizhevsky2012imagenet}. It contains 608 classes, which are separated into 351 classes for training, 97 for validation and 160 for testing.

\renewcommand{\tabcolsep}{3.5pt}
\begin{table*}[t]
\caption{Comparison with existing methods on 5-way classification task on benchmark miniImageNet and tieredImageNet datasets.
}
\centering
\begin{tabular}{ l | c | c c | c c}
\hline
\multicolumn{1}{c|}{\multirow{2}*{model}}  & \multirow{2}*{Backbone} & \multicolumn{2}{c|}{miniImageNet}  &\multicolumn{2}{c}{tieredImageNet} \\
\cline{3-6}
\multicolumn{1}{c|}{ } & & 1-shot &5-shot &1-shot &5-shot \\
\hline
MatchingNet \cite{vinyals2016matching} &Conv4 &43.44 $\pm$ 0.77 & 60.60 $\pm$ 0.71 & - & -\\
ProtoNet \cite{snell2017prototypical} &Conv4 &49.42 $\pm$ 0.78 & 68.20 $\pm$ 0.66 &53.31 $\pm$ 0.89 &72.69 $\pm$ 0.74\\
RelationNet \cite{sung2018learning} &Conv4 &50.44  $\pm$ 0.82 & 65.32  $\pm$ 0.70 & 54.48  $\pm$ 0.93 & 71.32  $\pm$ 0.78\\
\hdashline
\textbf{PatchProto} &Conv4 &54.71  $\pm$ 0.46 & 70.67  $\pm$ 0.38 & 56.90  $\pm$ 0.51 & 71.47  $\pm$ 0.42\\
\textbf{PatchProto+tSF}  &Conv4 &\textbf{57.39 $\pm$ 0.47} & \textbf{73.34  $\pm$ 0.37} & \textbf{59.79  $\pm$ 0.51} & \textbf{74.56  $\pm$ 0.41} \\
\hline
\hline
CAN \cite{hou2019cross} &ResNet-12 &63.85 $\pm$ 0.48 & 79.44 $\pm$ 0.34 &69.89 $\pm$ 0.51 &84.23 $\pm$ 0.37 \\
MetaOpt+ArL \cite{hongguang2021rethink} &ResNet-12 &65.21 $\pm$ 0.58 &80.41 $\pm$ 0.49 &- &-\\
DeepEMD \cite{zhang2020deepemd} &ResNet-12 &65.91 $\pm$ 0.82 & 82.41 $\pm$ 0.56 &71.16 $\pm$ 0.87 &86.03 $\pm$ 0.58 \\
IENet \cite{rizve2021exploring} &ResNet-12 &66.82 $\pm$ 0.80 & \textbf{84.35 $\pm$ 0.51} &71.87 $\pm$ 0.89 &\textbf{86.82 $\pm$ 0.58} \\
DANet \cite{xu2021learning} &ResNet-12 &67.76 $\pm$ 0.46 & 82.71 $\pm$ 0.31 &71.89 $\pm$ 0.52 &85.96 $\pm$ 0.35 \\
\hdashline
\textbf{PatchProto} &ResNet-12 &68.46  $\pm$ 0.47 & 82.65  $\pm$ 0.31 & 70.50  $\pm$ 0.50 & 83.60  $\pm$ 0.37 \\
\textbf{PatchProto+tSF}  &ResNet-12 &\textbf{69.74  $\pm$ 0.47} & 83.91  $\pm$ 0.30 & \textbf{71.98  $\pm$ 0.50} & 85.49  $\pm$ 0.35 \\
\hline
\hline
wDAE-GNN \cite{gidaris2019generating} &WRN-28 &61.07 $\pm$ 0.15 &76.75 $\pm$ 0.11 &68.18 $\pm$ 0.16 & 83.09 $\pm$ 0.12 \\
LEO \cite{rusu2019meta} &WRN-28 &61.76 $\pm$ 0.08 &77.59 $\pm$ 0.12 & 66.33 $\pm$ 0.05 & 81.44 $\pm$ 0.09 \\
PSST \cite{zhengyu2021pareto} &WRN-28 &64.16 $\pm$ 0.44 & 80.64 $\pm$ 0.32 &- &- \\
FEAT \cite{ye2020few} &WRN-28 &65.10 $\pm$ 0.20 & 81.11 $\pm$ 0.14 &70.41 $\pm$ 0.23 &84.38 $\pm$ 0.16 \\
CA \cite{afrasiyabi2019associative} &WRN-28 &65.92 $\pm$ 0.60 & 82.85 $\pm$ 0.55 &74.40 $\pm$ 0.68 &86.61 $\pm$ 0.59 \\
\hdashline
\textbf{PatchProto} &WRN-28 &69.34  $\pm$ 0.46 & 83.46  $\pm$ 0.30 & 73.40  $\pm$ 0.50 & 86.85  $\pm$ 0.35\\
\textbf{PatchProto+tSF}  &WRN-28 &\textbf{70.23  $\pm$ 0.46} & \textbf{84.55  $\pm$ 0.29} & \textbf{74.87  $\pm$ 0.49} & \textbf{88.05  $\pm$ 0.32} \\
\hline
\end{tabular}
\vspace{-0.4cm}
\label{table:SOTA}
\end{table*}

\noindent\textbf{Evaluation and Implementation details} \ \
We conduct experiments under $5$-way $1$-shot and $5$-shot settings. We report the \textit{average accuracy} and $95\%$ \textit{confidence interval} over $2000$ episodes sampled from the test set. Horizontal flipping, random cropping, random erasing \cite{zhong2017random} and color jittering are employed for data augmentation in training.
According to the ablation study results in Tab.~\ref{table:ablation_2}, the hyperparameter $\lambda$ in Eq.~\ref{equ:Loss} is set to $0.5$ and $1.5$ for ResNet-12 and WRN-28 respectively.
The detailed info of optimizer, learning-rate and training-epochs are referred to our public source code.

\noindent\textbf{Comparison with State-of-the-arts} \ \
Tab.~\ref{table:SOTA} compares our methods with existing few-shot classification algorithms on miniImageNet and tieredImageNet, which shows that the proposed PatchProto and PatchProto+tSF mthods outperform the existing SOTAs under different backbones. And PatchProto+tSF shows obviously accuracy improvements under different backbones on 1-shot and 5-shot tasks than the strong baseline PatchProto, which demonstrates the effectiveness of the proposed tSF. The proposed PatchProto+tSF performs better than the optimization-based MetaOpt+ArL \cite{hongguang2021rethink} and parameter-generating method wDAE-GCNN \cite{gidaris2019generating}, with an improvement up to $4.53\%$ and $9.16\%$ respectively.
Comparing to the competitive metric-based DeepEMD \cite{zhang2020deepemd}, PatchProto+tSF achieves $3.83\%$ higher accuracy. Some metric-based methods \cite{xu2021learning,hou2019cross} employing cross attention mechanism, and our PatchProto+tSF still surpasses the best DANet \cite{xu2021learning} with an performance improvement up to $1.98\%$ on 1-shot.
Overall our PatchProto+tSF obtains a new SOTA performance on both 1-shot and 5-shot classification tasks on miniImageNet and tieredImageNet, which demonstrates the strength of our framework and the effectiveness of the proposed tSF.

\renewcommand{\tabcolsep}{3.0pt}
\begin{table}[t]
\caption{The results on 5-way miniImageNet classification about the structure (refers to Fig.\ref{fig:tSF} (a) and Fig.\ref{fig:tSF} (b)) influence of tSF, which utilize PatchProto+tSF framework under ResNet-12 backbone. The dimension ${n}$ of ${\theta}$ in tSF is set to 5. The Metric and Global loss weights are set to 0.5 and 1.0 respectively, and the Rotation classifier is not applied.}
\vspace{-0.2cm}
\centering
\begin{tabular}{c | c | c | c | c  c}
\hline
\multirow{2}*{Neck} & \multirow{2}*{$Q,K,V$} & \multirow{1}*{Attention} & \multirow{2}*{Param} & \multicolumn{2}{c}{miniImageNet} \\
\cline{5-6}
 & & Heads & & 1-shot &5-shot \\
\hline
None  & - & - & 7.75M &67.47 $\pm$ 0.47 &81.85 $\pm$ 0.32  \\
\hdashline
Transformer  & ${Q}$=${K}$=${V}$=${f}$ &1 & 8.75M &63.25 $\pm$ 0.45 &79.44 $\pm$ 0.33  \\
Transformer  & ${Q}$=${W_Qf}$,${K}$=${W_Kf}$,${V}$=${W_Vf}$ &1 & 9.75M &62.96 $\pm$ 0.47 &78.92 $\pm$ 0.33  \\
Transformer  & ${Q}$=${K}$=${V}$=${f}$ &4 & 8.75M &62.68 $\pm$ 0.47 &78.98 $\pm$ 0.34  \\
Transformer  & ${Q}$=${W_Qf}$,${K}$=${W_Kf}$,${V}$=${W_Vf}$ &4 & 9.75M &62.70 $\pm$ 0.47 &78.33 $\pm$ 0.34  \\
\hdashline
DETR  & ${Q}$=${\theta}$,${K}$=${V}$=${f}$ &1 & 8.75M &63.55 $\pm$ 0.45 &79.65 $\pm$ 0.33  \\
\hdashline
tSF-V  & ${Q}$=${K}$=${f}$,${V}$=${\theta}$ &1 & 9.00M &61.84 $\pm$ 0.48 &76.11 $\pm$ 0.36  \\
tSF-K  & ${Q}$=${V}$=${f}$,${K}$=${\theta}$ &1 & 9.00M &64.73 $\pm$ 0.45 &80.43 $\pm$ 0.33  \\
\hdashline
tSF  & ${Q}$=${f}$,${K}$=${V}$=${\theta}$ &1 & 8.75M &68.37 $\pm$ 0.46 &83.08 $\pm$ 0.31  \\
tSF  & ${Q}$=${W_Qf}$,${K}$=${W_K\theta}$,${V}$=${W_V\theta}$ &1 & 9.75M &68.27 $\pm$ 0.47 &83.01 $\pm$ 0.31  \\
tSF  & ${Q}$=${f}$,${K}$=${V}$=${\theta}$ &4 & 8.75M &\textbf{68.60 $\pm$ 0.47} &\textbf{83.26 $\pm$ 0.31}  \\
tSF  & ${Q}$=${W_Qf}$,${K}$=${W_K\theta}$,${V}$=${W_V\theta}$ &4 & 9.75M &68.49 $\pm$ 0.47 &82.95 $\pm$ 0.31  \\
tSF  & ${Q}$=${f}$,${K}$=${V}$=${\theta}$ &8 & 8.75M &68.46 $\pm$ 0.46 &83.14 $\pm$ 0.31  \\
tSF  & ${Q}$=${W_Qf}$,${K}$=${W_K\theta}$,${V}$=${W_V\theta}$ &8 & 9.75M &68.42 $\pm$ 0.47 &83.12 $\pm$ 0.31  \\
\hline
\end{tabular}
\vspace{-0.2cm}
\label{table:ablation_1}
\end{table}

\begin{figure}[!t]
\centering
\includegraphics[width=0.99\linewidth]{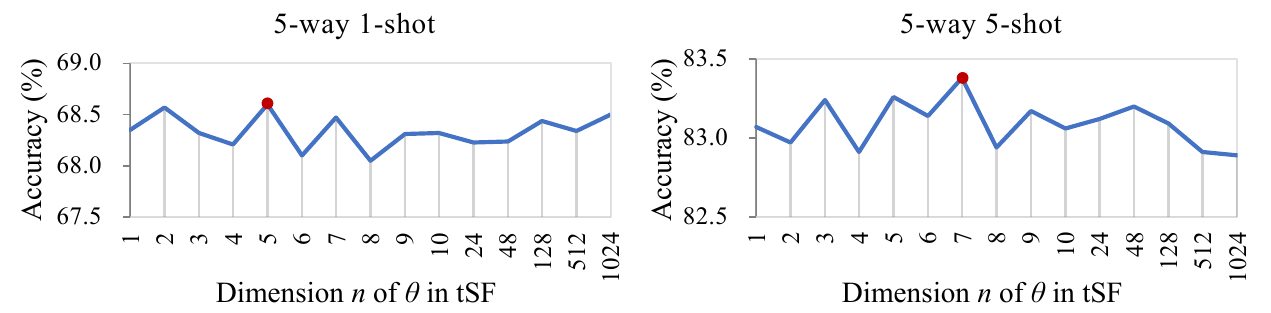}
\vspace{-1mm}
\caption{The results on miniImageNet classification about the influence of dimension ${n}$ of ${\theta \in \mathbb{R}^{n\times c}}$ in tSF, which utilize PatchProto+tSF framework under ResNet-12 backbone without Rotation classifier.}
\label{fig:dim}
\vspace{-4mm}
\end{figure}

\noindent\textbf{Ablation Study} \ \
\emph{Structure Influence of tSF:} As shown in Tab.~\ref{table:ablation_1}, by comparing our tSF to the baseline without neck in first row, it shows consistent improvements on 1-shot and 5-shot tasks, because tSF can transfer the knowledge from base set to novel set and generates more discriminative representations via focusing on the foreground regions.
Comparing to the baseline without neck, the self-attention based transformer shows performance degradation due to overfitting on base dataset.
Instead of behaving self-attention as standard transformer do, the proposed dataset-attention based tSF is able to prevent overfitting and generalize well on novel task, which is illustrated by the large accuracy improvement of tSF.

\renewcommand{\tabcolsep}{3.5pt}
\begin{table}[t]
\caption{The results on 5-way miniImageNet classification about the influence of multi-task loss employed in PatchProto+tSF under ResNet-12 and WRN-28 backbones.
As introduced in Eq.~\ref{equ:Loss}, ${\lambda}$ is the hyper-parameter, and ${w_G}, {w_R}$ are learnable weights.}
\centering
\begin{tabular}{c | c c c | c  c | c  c}
\hline
\multirow{2}*{${\lambda}$} & \multicolumn{3}{c|}{Loss weights} & \multicolumn{2}{c|}{ResNet-12} & \multicolumn{2}{c}{WRN-28} \\
\cline{2-8}
&Metric &Global &Rotation & 1-shot &5-shot & 1-shot &5-shot \\
\hline
-&0.5 &- &- &62.76 $\pm$ 0.49 &80.07 $\pm$ 0.34 &61.71 $\pm$ 0.50 &77.53 $\pm$ 0.36 \\
-&0.5 &- &1.0 &65.57 $\pm$ 0.49 &80.81 $\pm$ 0.34 &63.97 $\pm$ 0.50 &79.25 $\pm$ 0.37 \\
-&0.5 &1.0 &- &68.60 $\pm$ 0.47 &82.98 $\pm$ 0.31 &67.73 $\pm$ 0.47 &82.59 $\pm$ 0.31 \\
-&0.5 &1.0 &1.0 &\textbf{69.41 $\pm$ 0.46} &\textbf{83.82 $\pm$ 0.30} &\textbf{69.64 $\pm$ 0.47} &\textbf{84.01 $\pm$ 0.30} \\
\hline
0.0&0.5 &${\lambda}$+${w_G}$ &${\lambda}$+${w_R}$ &68.19 $\pm$ 0.47 &83.00 $\pm$ 0.31 &68.55 $\pm$ 0.48 &83.30 $\pm$ 0.31  \\
0.5&0.5 &${\lambda}$+${w_G}$ &${\lambda}$+${w_R}$ &\textbf{69.74 $\pm$ 0.47} &\textbf{83.91 $\pm$ 0.30} &69.20 $\pm$ 0.46 &84.03 $\pm$ 0.30 \\
1.0&0.5 &${\lambda}$+${w_G}$ &${\lambda}$+${w_R}$ &69.44 $\pm$ 0.46 &83.90 $\pm$ 0.31 &70.05 $\pm$ 0.46 &84.17 $\pm$ 0.29  \\
1.5&0.5 &${\lambda}$+${w_G}$ &${\lambda}$+${w_R}$ &69.30 $\pm$ 0.45 &83.82 $\pm$ 0.30 &\textbf{70.23 $\pm$ 0.46} &\textbf{84.55 $\pm$ 0.29}  \\
2.0&0.5 &${\lambda}$+${w_G}$ &${\lambda}$+${w_R}$ &69.50 $\pm$ 0.45 &83.86 $\pm$ 0.30 &70.02 $\pm$ 0.45 &83.61 $\pm$ 0.30 \\
\hline
\end{tabular}
\vspace{-0.2cm}
\label{table:ablation_2}
\end{table}

\renewcommand{\tabcolsep}{1.3pt}
\begin{table*}[t!]
\centering
\small
\caption{The results on 1-way PASCAL-5$^i$ segmentation using mean-IoU. The dimension ${n}$ of ${\theta}$ in tSF is set to 5.}
{
\begin{tabular}{lcccccacccca}
    \toprule
     & & \multicolumn{5}{c}{1 shot} & \multicolumn{5}{c}{5 shot} \\
     \cmidrule(lr){3-7}\cmidrule(lr){8-12}
     Method & Backbone & Fold-0 & Fold-1 & Fold-2 & Fold-3 & Mean & Fold-0 & Fold-1 & Fold-2 & Fold-3 & Mean \\
     \midrule
     PANet \cite{wang2019panet} & \multirow{2}{*}{VGG-16} &42.3 & 58.0 & 51.1 & 41.2 & 48.1 & 51.8 & 64.6 & 59.8 & 46.5 & 55.7  \\
     RPMM \cite{ppnet} & & 47.1 & \textbf{65.8} & 50.6 & \textbf{48.5} & 53.0 & 50.0&  66.5 & 51.9 & 47.6 & 54.0 \\
     \hdashline
     \textbf{RePRI} \cite{Malik2021repri}& \multirow{2}{*}{VGG-16} &  49.7 & 63.4 & 58.2 & 42.8 & 53.5 & 54.5 & 67.2 & 63.7 & 48.8 & 58.6 \\
     \textbf{RePRI+tSF} & &  \textbf{53.0} & {65.3} & \textbf{58.3} & {44.2} & \textbf{55.2} & \textbf{57.0} & \textbf{67.9} & \textbf{63.9} & \textbf{50.8} & \textbf{59.9} \\
     \hline
     CANet \cite{zhang2019canet} & \multirow{4}{*}{ResNet-50} & 52.5 & 65.9 & 51.3 & \textbf{51.9} &  55.4 & 55.5 & 67.8 & 51.9 & 53.2  & 57.1 \\
     PGNet \cite{zhang2019pyramid}  &  &  56.0 & 66.9 & 50.6 & 50.4 &  56.0 & 57.7 & 68.7 & 52.9 & 54.6  & 58.5 \\

     RPMM \cite{rpmm} & &55.2 & 66.9 & 52.6 & 50.7 & 56.3 & 56.3 & 67.3 &  54.5 & 51.0 & 57.3  \\
     PPNet \cite{ppnet} & &47.8 & 58.8 & 53.8 & 45.6 & 51.5 & 58.4 & 67.8 &  64.9 &56.7 &  62.0  \\
     \hdashline
     \textbf{RePRI} \cite{Malik2021repri} & \multirow{2}{*}{ResNet-50}&  60.8 & 67.8 & 60.9 & 47.5 & 59.3 & 66.0 & 70.9 & 65.9 & 56.4 & 64.8 \\
     \textbf{RePRI+tSF} & &  \textbf{62.4} & \textbf{68.6} & \textbf{61.4} & {49.4} & \textbf{60.5} & \textbf{66.4} & \textbf{71.1} & \textbf{66.4} & \textbf{58.3} & \textbf{65.6} \\
    \bottomrule\\
\end{tabular}
}
\label{tab:FSSS_PASCAL_results}
\vspace{-8mm}
\end{table*}

\renewcommand{\tabcolsep}{1.3pt}
\begin{table*}[h]
\centering
\small
\caption{The results on 1-way COCO-20$^i$ segmentation using mean-IoU. The dimension ${n}$ of ${\theta}$ in tSF is set to 5.}
{
\begin{tabular}{lcccccacccca}
\toprule
& & \multicolumn{5}{c}{1 shot} & \multicolumn{5}{c}{5 shot} \\
\cmidrule(lr){3-7}\cmidrule(lr){8-12}
Method & Backbone & Fold-0 & Fold-1 & Fold-2 & Fold-3 & Mean & Fold-0 & Fold-1 & Fold-2 & Fold-3 & Mean \\
\midrule
PPNet \cite{ppnet} & \multirow{3}{*}{ResNet-50} & 34.5 & 25.4 & 24.3 & 18.6 & 25.7 & \textbf{48.3} & 30.9 & 35.7 & 30.2 & 36.2\\
RPMM \cite{rpmm}  &  & 29.5 & 36.8 & 29.0 & 27.0 & 30.6 & 33.8 & 42.0 & 33.0 & 33.3 &  35.5\\
PFENet \cite{pfenet} &  &  {36.5} & {38.6} & {34.5} & {33.8} & {35.8} & 36.5 & 43.3 & 37.8 & 38.4 & 39.0 \\
\hdashline
\textbf{RePRI} \cite{Malik2021repri} & \multirow{2}{*}{ResNet-50} & 36.1 & 40.0 & 34.0 & 36.1 & 36.6 & {43.3} & {48.7} & {44.0} & {44.9} &{45.2} \\
\textbf{RePRI+tSF} &  & \textbf{38.4} & \textbf{41.3} & \textbf{35.2} & \textbf{37.7} & \textbf{38.2} & {45.0} & \textbf{49.9} & \textbf{45.5} & \textbf{45.6} & \textbf{46.5} \\
\bottomrule
\end{tabular}
}
\label{tab:FSSS_COCO_results}
\end{table*}

\begin{table}[]
\centering
{
\caption{The results on VOC dataset. we evaluate the performance($AP_{50}$) of DeFRCN under ResNet-101 with tSF module on three novel splits over multiple runs. The term \textit{w/G} denotes whether using \textit{G-FSOD} setting \cite{wang2020frustratingly}. The dimension ${n}$ of ${\theta}$ in tSF is set to 5.
}
\label{tab:voc_result}
\resizebox{\textwidth}{!}
{
\begin{tabular}{c|l|l|l|c|ccccc|ccccc|ccccc}
\toprule[1.1pt]
\multicolumn{4}{c|}{}            & \multicolumn{1}{c|}{}                    & \multicolumn{5}{c|}{Novel Set 1}                                              & \multicolumn{5}{c|}{Novel Set 2}                                              & \multicolumn{5}{c}{Novel Set 3}                                              \\
\multicolumn{4}{c|}{\multirow{-2}{*}{Method / Shots}} &\multicolumn{1}{c|}{\multirow{-2}{*}{\textit{w/G}}} & 1             & 2             & 3             & 5             & 10            & 1             & 2             & 3             & 5             & 10            & 1             & 2             & 3             & 5             & 10            \\ \midrule[0.9pt]
\multicolumn{4}{c|}{MetaDet \cite{wang2019meta}}     & \XSolidBrush           & 18.9          & 20.6          & 30.2          & 36.8          & 49.6          & {21.8}          & 23.1          & 27.8          & 31.7          & 43.0          & 20.6          & 23.9          & 29.4          & 43.9          & 44.1          \\
\multicolumn{4}{c|}{TFA \cite{wang2020frustratingly}} & \XSolidBrush             & 39.8          & 36.1        & 44.7          & 55.7         & 56.0          & 23.5          & 26.9         & 34.1          & 35.1          & 39.1          & 30.8          & 34.8          & 42.8        & 49.5         & 49.8       \\
\midrule[0.9pt]
\rowcolor[HTML]{EFEFEF}
	\multicolumn{4}{c|}{\textbf{DeFRCN} \cite{qiao2021defrcn}} & \XSolidBrush   & \textbf{53.6} & 57.5 & \textbf{61.5} & \textbf{64.1} & \textbf{60.8} & 30.1 & 38.1 & \textbf{47.0} & \textbf{53.3} & \textbf{47.9} & 48.4 & \textbf{50.9} & 52.3 & \textbf{54.9} & 57.4 \\
\rowcolor[HTML]{EFEFEF}
	\multicolumn{4}{c|}{\textbf{DeFRCN+tSF}} & \XSolidBrush   & \textbf{53.6} & \textbf{58.1} & \textbf{61.5} & 63.8 & \textbf{60.8} & \textbf{31.5} & \textbf{39.3} & \textbf{47.0} & 52.1 & 47.3 & \textbf{48.5} & 50.5 & \textbf{52.8} & 54.5 & \textbf{58.0} \\
\midrule[0.9pt]


\multicolumn{4}{c|}{TFA \cite{wang2020frustratingly}}     & \CheckmarkBold         & 25.3         & 36.4          & 42.1          & 47.9          & 52.8          & 18.3          & 27.          & 30.9          & 34.1          & 39.5          & 17.9          & 27.2          & 34.3          & 40.8          & 45.6          \\
\multicolumn{4}{c|}{FSDetView \cite{Xiao2020FSDetView}}        &\CheckmarkBold      & 24.2          & 35.3          &  42.2         & 49.1          & 57.4       & 21.6         & 24.6          & 31.9       & 37.0         & 45.7         & 21.2         & 30.0         & 37.2         & 43.8         & 49.6  \\
\midrule[0.9pt]
\rowcolor[HTML]{EFEFEF}
\multicolumn{4}{c|}{\textbf{DeFRCN} \cite{qiao2021defrcn}}   & \CheckmarkBold  & 40.2 & 53.6 & 58.2 & 63.6 & \textbf{66.5} & 29.5 & 39.7 & 43.4 & 48.1 & \textbf{52.8} & 35.0 & 38.3 & 52.9 & 57.7 & 60.8 \\
\rowcolor[HTML]{EFEFEF}
\multicolumn{4}{c|}{\textbf{DeFRCN+tSF}}   & \CheckmarkBold  & \textbf{43.6} & \textbf{57.4} & \textbf{61.2} & \textbf{65.1} & 65.9 & \textbf{31.0} & \textbf{40.3} & \textbf{45.3} & \textbf{49.6} & 52.5 & \textbf{39.3} & \textbf{51.4} & \textbf{54.8} & \textbf{59.8} & \textbf{62.1} \\

\bottomrule[1.1pt]
\end{tabular}}}
\vspace{-0.35cm}
\end{table}

\noindent\emph{Dimension Influence of ${\theta}$ in tSF:} As shown in Fig.\ref{fig:dim}, with a wide range ${[1,1024]}$ dimension ${n}$ of ${\theta \in \mathbb{R}^{n\times c}}$ in tSF, the accuracy differences are within ${0.5\%}$ on 1-shot and 5-shot tasks, i.e. our PatchProto+tSF framework is not sensitive to the dimension of ${\theta}$. Considering the accuracy performance and computational complexity, we recommend to set the dimension ${n=5}$.
\ljx{The $n$ is interpreted as number of semantic groups. As $n$ going larger, semantic groups become more fine-grained. The setting of ${n=1}$ represents one foreground group and is still able to obtain impressive performance.}

\noindent\emph{Influence of multi-task loss:}
As illustrated in Tab.~\ref{table:ablation_2}, the proposed PatchProto+tSF obtains its best results as setting ${\lambda}$ to ${0.5}$ and ${1.5}$ under ResNet-12 and WRN-28 backbones respectively.
Comparing to the baseline (i.e. with Metric classification task only) in first row, our multi-task framework achieves large improvements on 1-shot and 5-shot tasks under different backbones.
These results indicate that the auxiliary tasks (i.e. Global classification and Rotation classification) are useful for training a more robust embedding leading to an accuracy improvement, and the weights of the auxiliary tasks have a great influence on the overall few-shot classification performance.

\subsection{Few-Shot Semantic Segmentation}
\noindent\textbf{Datasets and Setting} \ \
\emph{PASCAL-5$^i$ and COCO-20$^i$ Datasets:}
\ljx{
PASCAL-5$^i$ is built from PASCAL VOC \cite{everingham2010pascal}.
The 20 object categories are split into 4 folds. For each fold, 15 categories are utilized for training and the remaining 5 classes for testing.
COCO-20$^i$ is built from MS-COCO \cite{ms-COCO}. COCO-20$^i$ dataset is divided into 4 folds with 60 base classes and 20 test classes in each fold.
}

\noindent\emph{Evaluation Setting:}
Following \cite{ppnet}, the mean Intersection over Union (mIoU) is adopted for evaluation, and we report the average mIoU over 5 runs of 1000 tasks.

\noindent\textbf{Comparison with State-of-the-arts} \ \
Tab.~\ref{tab:FSSS_PASCAL_results} and Tab.~\ref{tab:FSSS_COCO_results} present our evaluation results on PASCAL-5$^i$ and COCO-20$^i$. Comparing with existing few-shot semantic segmentation methods, the RePRI+tSF approach achieves new state-of-the-art results.
With the help of our tSF module, the RePRI+tSF obtains consistent performance improvement than RePRI, on 1-way 1-shot and 5-shot tasks under VGG-16 and ResNet-50 backbones.

\subsection{Few-Shot Object Detection}
\noindent\textbf{Datasets and Setting} \ \
\emph{PASCAL VOC and COCO Datasets:}
PASCAL VOC \cite{everingham2010pascal} are randomly sampled into 3 splits, and each contains 20 categories. For each split, there are 15 base and 5 novel categories.
Each novel class has $K = 1,2,3,5,10$ objects sampled from the train/val set of VOC2007 and VOC2012 for training, and the test set of VOC2007 for testing.
COCO \cite{lin2014microsoft} use 60 categories disjoint with VOC as base set, and the remaining 20 categories are novel set with $K = 1,2,3,5,10,30$ shots. The total 5k images randomly sampled from the validation set are utilized for testing, while the rest for training.

\renewcommand{\tabcolsep}{1.5pt}
\begin{wraptable}{r}{0.52\textwidth}
\vspace{-0.9cm}
\centering
\caption{The results on COCO dataset. we report the performance ($mAP$) of DeFRCN under ResNet-101 with tSF module over multiple runs. The dimension ${n}$ of ${\theta}$ in tSF is set to 5.}
\vspace{0.2cm}
\label{tab:coco_result}
\resizebox{0.52\textwidth}{!}
{\begin{tabular}{c|l|l|l|c|cccccc}
\toprule[1.0pt]
\multicolumn{4}{c|}{}            & \multicolumn{1}{c|}{}                    & \multicolumn{6}{c}{Shot Number}                                       \\
\multicolumn{4}{c|}{\multirow{-2}{*}{Method / Shots}} &\multicolumn{1}{c|}{\multirow{-2}{*}{\textit{w/G}}} & 1        & 2              & 3             & 5             & 10    &30                \\ \midrule[0.9pt]
\multicolumn{4}{c|}{TFA \cite{wang2020frustratingly}} & \XSolidBrush             & 4.4  &5.4       &6.0          & 7.7         &10.0         & 13.7                 \\
\multicolumn{4}{c|}{FSDetView \cite{Xiao2020FSDetView}}        & \XSolidBrush      &4.5  &6.6       &7.2          & 10.7          & 12.5        & 14.7          \\
\midrule[0.9pt]
\rowcolor[HTML]{EFEFEF}
\multicolumn{4}{c|}{\textbf{DeFRCN} \cite{qiao2021defrcn}} & \XSolidBrush   & 9.3 & 12.9 & \textbf{14.8} & 16.1 & \textbf{18.5} & \textbf{22.6} \\
\rowcolor[HTML]{EFEFEF}
\multicolumn{4}{c|}{\textbf{DeFRCN+tSF}} & \XSolidBrush   & \textbf{9.9} & \textbf{13.5} & \textbf{14.8} & \textbf{16.3} & 18.3 & 22.5 \\
\midrule[0.9pt]

\multicolumn{4}{c|}{TFA \cite{wang2020frustratingly}}     & \CheckmarkBold         & 1.9         & 3.9  & 5.1          & 7.0          & 9.1          & 12.1                 \\
\multicolumn{4}{c|}{FSDetView \cite{Xiao2020FSDetView}}        & \CheckmarkBold      & 3.2         & 4.9  & 6.7         & 8.1         & 10.7        & 15.9           \\
\midrule[0.9pt]
\rowcolor[HTML]{EFEFEF}
\multicolumn{4}{c|}{\textbf{DeFRCN} \cite{qiao2021defrcn}}   & \CheckmarkBold  & 4.8 & 8.5 & 10.7 & \textbf{13.6} & \textbf{16.8} & \textbf{21.2}  \\
\rowcolor[HTML]{EFEFEF}
\multicolumn{4}{c|}{\textbf{DeFRCN+tSF}}   & \CheckmarkBold  & \textbf{5.0} & \textbf{8.7} & \textbf{10.9} & \textbf{13.6} & 16.6 & 20.9  \\
\bottomrule[1.0pt]
\end{tabular}}
\vspace{-0.4cm}
\end{wraptable}

\noindent\emph{Evaluation Setting:} Following \cite{wang2020frustratingly,Xiao2020FSDetView,kang2019few,yan2019meta,qiao2021defrcn}, we conduct experiments on two evaluation protocols: few-shot object detection \textit{\textbf{(FSOD)}} and generalized few-shot object detection \textit{\textbf{(G-FSOD)}}.
The setting of FSOD only observes the performance of novel set.
More comprehensively, that of G-FSOD considers both novel set and base set.

\noindent\textbf{Comparison with State-of-the-arts} \ \  Tab.~\ref{tab:voc_result} and Tab.~\ref{tab:coco_result} present our evaluation results on VOC and COCO. Comparing with existing few-shot object detection methods, the DeFRCN+tSF approach achieves new state-of-the-art on VOC and COCO. On VOC three different data splits, under the FSOD and G-FSOD setting, the indicators of all shots have been improved to a certain extent afer adding tSF module. We achieve around $2.6AP$ improvement over the DeFRCN in all shots under G-FSOD setting. On COCO dataset, DeFRCN+tSF consistently outperforms DeFRCN in 1, 2, 3 and 5 shots.

\section{Conclusions}
In this paper, we propose a transformer-based semantic filter (tSF) for few-shot learning problem.
tSF leverages a well-designed transformer-based structure to encode the knowledge from base dataset and novel samples.
In this way, a target-aware and image-aware feature can be generated via tSF.
Moreover, tSF is a universal module, which can be applied into multiple few-shot learning tasks, e.g., classification, segmentation and detection.
In addition, the parameter size of tSF is half of that of a standard transformer (less than $1M$).
The experimental results show that tSF is able to improve the performances about $2\%$ in multiple classic few-shot learning tasks.



%
%
\bibliographystyle{splncs04}
\bibliography{egbib}

\begin{thebibliography}{10}
\providecommand{\url}[1]{\texttt{#1}}
\providecommand{\urlprefix}{URL }
\providecommand{\doi}[1]{https://doi.org/#1}

\bibitem{afrasiyabi2019associative}
Afrasiyabi, A., Lalonde, J.F., Gagn{\'e}, C.: Associative alignment for
  few-shot image classification. In: ECCV (2020)

\bibitem{marcin2018learn}
Andrychowicz, M., Denil, M., Gomez, S., Hoffman, M.W., Pfau, D., Schaul, T.,
  Shillingford, B., De~Freitas, N.: Learning to learn by gradient descent by
  gradient descent. In: NeurIPS (2016)

\bibitem{bertinetto2016learning}
Bertinetto, L., Henriques, J.F., Valmadre, J., Torr, P., Vedaldi, A.: Learning
  feed-forward one-shot learners. In: NeurIPS (2016)

\bibitem{carion2020end}
Carion, N., Massa, F., Synnaeve, G., Usunier, N., Kirillov, A., Zagoruyko, S.:
  End-to-end object detection with transformers. In: ECCV (2020)

\bibitem{devlin2018bert}
Devlin, J., Chang, M.W., Lee, K., Toutanova, K.: Bert: Pre-training of deep
  bidirectional transformers for language understanding. In: NAACL-HLT (2020)

\bibitem{doersch2020crosstransformers}
Doersch, C., Gupta, A., Zisserman, A.: Crosstransformers: spatially-aware
  few-shot transfer. In: NeurIPS (2020)

\bibitem{dong2018few}
Dong, N., Xing, E.: Few-shot semantic segmentation with prototype learning. In:
  British Machine Vision Conference (BMVC) (2018)

\bibitem{dosovitskiy2020image}
Dosovitskiy, A., Beyer, L., Kolesnikov, A., Weissenborn, D., Zhai, X.,
  Unterthiner, T., Dehghani, M., Minderer, M., Heigold, G., Gelly, S., et~al.:
  An image is worth 16x16 words: Transformers for image recognition at scale.
  In: ICLR (2021)

\bibitem{everingham2010pascal}
Everingham, M., Van~Gool, L., Williams, C.K., Winn, J., Zisserman, A.: The
  pascal visual object classes (voc) challenge. International journal of
  computer vision  \textbf{88}(2),  303--338 (2010)

\bibitem{fan2020few}
Fan, Q., Zhuo, W., Tang, C.K., Tai, Y.W.: Few-shot object detection with
  attention-rpn and multi-relation detector. In: CVPR (2020)

\bibitem{fan2021generalized}
Fan, Z., Ma, Y., Li, Z., Sun, J.: Generalized few-shot object detection without
  forgetting. In: CVPR (2021)

\bibitem{finn2017model}
Finn, C., Abbeel, P., Levine, S.: Model-agnostic meta-learning for fast
  adaptation of deep networks. In: ICML (2017)

\bibitem{gidaris2019boosting}
Gidaris, S., Bursuc, A., Komodakis, N., P{\'e}rez, P., Cord, M.: Boosting
  few-shot visual learning with self-supervision. In: ICCV (2019)

\bibitem{gidaris2019generating}
Gidaris, S., Komodakis, N.: Generating classification weights with gnn
  denoising autoencoders for few-shot learning. In: CVPR (2019)

\bibitem{gregory2015siamese}
Gregory, K., Richard, Z., Ruslan, S.: Siamese neural networks for one-shot
  image recognition. In: ICML Workshops (2015)

\bibitem{hongguang2021rethink}
Hongguang, Z., Piotr, K., Songlei, J., Hongdong, L., Philip, H.~S., T.:
  Rethinking class relations: Absolute-relative supervised and unsupervised
  few-shot learning. In: CVPR (2021)

\bibitem{hou2019cross}
Hou, R., Chang, H., Bingpeng, M., Shan, S., Chen, X.: Cross attention network
  for few-shot classification. In: NeurIPS (2019)

\bibitem{hu2021dense}
Hu, H., Bai, S., Li, A., Cui, J., Wang, L.: Dense relation distillation with
  context-aware aggregation for few-shot object detection. In: CVPR (2021)

\bibitem{kang2019few}
Kang, B., Liu, Z., Wang, X., Yu, F., Feng, J., Darrell, T.: Few-shot object
  detection via feature reweighting. In: ICCV (2019)

\bibitem{krizhevsky2012imagenet}
Krizhevsky, A., Sutskever, I., Hinton, G.E.: Imagenet classification with deep
  convolutional neural networks. In: NeurIPS (2012)

\bibitem{lai2022rethinking}
Lai, J., Yang, S., Jiang, G., Wang, X., Li, Y., Jia, Z., Chen, X., Liu, J.,
  Gao, B.B., Zhang, W., et~al.: Rethinking the metric in few-shot learning:
  From an adaptive multi-distance perspective. In: Proceedings of the 30th ACM
  International Conference on Multimedia. pp. 4021--4030 (2022)

\bibitem{liang2020polytransform}
Liang, J., Homayounfar, N., Ma, W.C., Xiong, Y., Hu, R., Urtasun, R.:
  Polytransform: Deep polygon transformer for instance segmentation. In: CVPR
  (2020)

\bibitem{lin2014microsoft}
Lin, T.Y., Maire, M., Belongie, S., Hays, J., Perona, P., Ramanan, D.,
  Doll{\'a}r, P., Zitnick, C.L.: Microsoft coco: Common objects in context. In:
  ECCV (2014)

\bibitem{ms-COCO}
Lin, T.Y., Maire, M., Belongie, S., Hays, J., Perona, P., Ramanan, D.,
  Doll{\'a}r, P., Zitnick, C.L.: Microsoft coco: Common objects in context. In:
  ECCV (2014)

\bibitem{liu2021learning}
Liu, C., Fu, Y., Xu, C., Yang, S., Li, J., Wang, C., Zhang, L.: Learning a
  few-shot embedding model with contrastive learning. In: AAAI (2021)

\bibitem{liu2020crnet}
Liu, W., Zhang, C., Lin, G., Liu, F.: C{RN}et: Cross-reference networks for
  few-shot segmentation. In: CVPR (2020)

\bibitem{ppnet}
Liu, Y., Zhang, X., Zhang, S., He, X.: Part-aware prototype network for
  few-shot semantic segmentation. In: ECCV (2020)

\bibitem{lu2021simpler}
Lu, Z., He, S., Zhu, X., Zhang, L., Song, Y.Z., Xiang, T.: Simpler is better:
  Few-shot semantic segmentation with classifier weight transformer. In: ICCV
  (2021)

\bibitem{Malik2021repri}
Malik, B., Hoel, K., Ziko, I.M., Pablo, P., Ismail, B.A., Jose, D.: Few-shot
  segmentation without meta-learning: A good transductive inference is all you
  need? In: CVPR (2021)

\bibitem{carlos2020self}
Medina, C., Devos, A., Grossglauser, M.: Self-supervised prototypical transfer
  learning for few-shot classification. arXiv:2006.11325  (2020)

\bibitem{min2021hypercorrelation}
Min, J., Kang, D., Cho, M.: Hypercorrelation squeeze for few-shot segmentation.
  In: ICCV (2021)

\bibitem{munkhdalai2017meta}
Munkhdalai, T., Yu, H.: Meta networks. In: ICML (2017)

\bibitem{munkhdalai2018rapid}
Munkhdalai, T., Yuan, X., Mehri, S., Trischler, A.: Rapid adaptation with
  conditionally shifted neurons. In: ICML (2018)

\bibitem{nguyen2019feature}
Nguyen, K., Todorovic, S.: Feature weighting and boosting for few-shot
  segmentation. In: ICCV (2019)

\bibitem{nichol2018first}
Nichol, A., Achiam, J., Schulman, J.: On first-order meta-learning algorithms.
  arXiv:1803.02999  (2018)

\bibitem{qi2018memory}
Qi, C., Yingwei, P., Ting, Y., Chenggang, Y., Tao, M.: Memory matching networks
  for one-shot image recognition. In: CVPR (2018)

\bibitem{qiao2021defrcn}
Qiao, L., Zhao, Y., Li, Z., Qiu, X., Wu, J., Zhang, C.: Defrcn: Decoupled
  faster r-cnn for few-shot object detection. In: ICCV (2021)

\bibitem{rakelly2018conditional}
Rakelly, K., Shelhamer, E., Darrell, T., Efros, A., Levine, S.: Conditional
  networks for few-shot semantic segmentation. In: ICLR Workshop (2018)

\bibitem{ravi2016optimization}
Ravi, S., Larochelle, H.: Optimization as a model for few-shot learning. In:
  ICLR (2017)

\bibitem{ren2018meta}
Ren, M., Triantafillou, E., Ravi, S., Snell, J., Swersky, K., Tenenbaum, J.B.,
  Larochelle, H., Zemel, R.S.: Meta-learning for semi-supervised few-shot
  classification. In: ICLR (2018)

\bibitem{rizve2021exploring}
Rizve, M.N., Khan, S., Khan, F.S., Shah, M.: Exploring complementary strengths
  of invariant and equivariant representations for few-shot learning. In: CVPR
  (2021)

\bibitem{rusu2019meta}
Rusu, A.A., Rao, D., Sygnowski, J., Vinyals, O., Pascanu, R., Osindero, S.,
  Hadsell, R.: Meta-learning with latent embedding optimization. In: ICLR
  (2019)

\bibitem{sarlin2020superglue}
Sarlin, P.E., DeTone, D., Malisiewicz, T., Rabinovich, A.: Superglue: Learning
  feature matching with graph neural networks. In: CVPR (2020)

\bibitem{shaban2017one}
Shaban, A., Bansal, S., Liu, Z., Essa, I., Boots, B.: One-shot learning for
  semantic segmentation. In: BMVC (2018)

\bibitem{siam2019amp}
Siam, M., Oreshkin, B.N., Jagersand, M.: A{MP}: Adaptive masked proxies for
  few-shot segmentation. In: ICCV (2019)

\bibitem{snell2017prototypical}
Snell, J., Swersky, K., Zemel, R.: Prototypical networks for few-shot learning.
  In: NeurIPS (2017)

\bibitem{su2020does}
Su, J.C., Maji, S., Hariharan, B.: When does self-supervision improve few-shot
  learning? In: ECCV (2020)

\bibitem{sun2021fsce}
Sun, B., Li, B., Cai, S., Yuan, Y., Zhang, C.: Fsce: Few-shot object detection
  via contrastive proposal encoding. In: CVPR (2021)

\bibitem{sun2021loftr}
Sun, J., Shen, Z., Wang, Y., Bao, H., Zhou, X.: Loftr: Detector-free local
  feature matching with transformers. In: CVPR (2021)

\bibitem{sung2018learning}
Sung, F., Yang, Y., Zhang, L., Xiang, T., Torr, P.H., Hospedales, T.M.:
  Learning to compare: Relation network for few-shot learning. In: CVPR (2018)

\bibitem{tian2020rethinking}
Tian, Y., Wang, Y., Krishnan, D., Tenenbaum, J.B., Isola, P.: Rethinking
  few-shot image classification: a good embedding is all you need? In: ECCV
  (2020)

\bibitem{pfenet}
Tian, Z., Zhao, H., Shu, M., Yang, Z., Li, R., Jia, J.: Prior guided feature
  enrichment network for few-shot segmentation. TPAMI  (2020)

\bibitem{touvron2021training}
Touvron, H., Cord, M., Douze, M., Massa, F., Sablayrolles, A., J{\'e}gou, H.:
  Training data-efficient image transformers \& distillation through attention.
  In: ICML (2021)

\bibitem{vaswani2017attention}
Vaswani, A., Shazeer, N., Parmar, N., Uszkoreit, J., Jones, L., Gomez, A.N.,
  Kaiser, {\L}., Polosukhin, I.: Attention is all you need. In: NeurIPS (2017)

\bibitem{vinyals2016matching}
Vinyals, O., Blundell, C., Lillicrap, T., Wierstra, D., et~al.: Matching
  networks for one shot learning. In: NeurIPS (2016)

\bibitem{wangfew}
Wang, H., Zhang, X., Hu, Y., Yang, Y., Cao, X., Zhen, X.: Few-shot semantic
  segmentation with democratic attention networks. In: ECCV (2020)

\bibitem{wang2019panet}
Wang, K., Liew, J.H., Zou, Y., Zhou, D., Feng, J.: Panet: Few-shot image
  semantic segmentation with prototype alignment. In: ICCV (2019)

\bibitem{wang2021pyramid}
Wang, W., Xie, E., Li, X., Fan, D.P., Song, K., Liang, D., Lu, T., Luo, P.,
  Shao, L.: Pyramid vision transformer: A versatile backbone for dense
  prediction without convolutions. In: ICCV (2021)

\bibitem{wang2020frustratingly}
Wang, X., Huang, T.E., Darrell, T., Gonzalez, J.E., Yu, F.: Frustratingly
  simple few-shot object detection. arXiv:2003.06957  (2020)

\bibitem{yan2019simpleshot}
Wang, Y., Chao, W.L., Weinberger, K.Q., van~der Maaten, L.: Simpleshot:
  Revisiting nearest-neighbor classification for few-shot learning.
  arXiv:1911.04623  (2019)

\bibitem{wang2019meta}
Wang, Y.X., Ramanan, D., Hebert, M.: Meta-learning to detect rare objects. In:
  ICCV (2019)

\bibitem{wu2020multi}
Wu, J., Liu, S., Huang, D., Wang, Y.: Multi-scale positive sample refinement
  for few-shot object detection. In: ECCV (2020)

\bibitem{wu2019parn}
Wu, Z., Li, Y., Guo, L., Jia, K.: Parn: Position-aware relation networks for
  few-shot learning. In: ICCV (2019)

\bibitem{xiao2020few}
Xiao, Y., Marlet, R.: Few-shot object detection and viewpoint estimation for
  objects in the wild. In: ECCV (2020)

\bibitem{Xiao2020FSDetView}
Xiao, Y., Marlet, R.: Few-shot object detection and viewpoint estimation for
  objects in the wild. In: ECCV (2020)

\bibitem{xie2021segformer}
Xie, E., Wang, W., Yu, Z., Anandkumar, A., Alvarez, J.M., Luo, P.: Segformer:
  Simple and efficient design for semantic segmentation with transformers. In:
  NeurIPS (2021)

\bibitem{xu2021learning}
Xu, C., Fu, Y., Liu, C., Wang, C., Li, J., Huang, F., Zhang, L., Xue, X.:
  Learning dynamic alignment via meta-filter for few-shot learning. In: CVPR
  (2021)

\bibitem{yan2019meta}
Yan, X., Chen, Z., Xu, A., Wang, X., Liang, X., Lin, L.: Meta r-cnn: Towards
  general solver for instance-level low-shot learning. In: ICCV (2019)

\bibitem{rpmm}
Yang, B., Liu, C., Li, B., Jiao, J., Ye, Q.: Prototype mixture models for
  few-shot semantic segmentation. In: ECCV (2020)

\bibitem{yang2020context}
Yang, Z., Wang, Y., Chen, X., Liu, J., Qiao, Y.: Context-transformer: tackling
  object confusion for few-shot detection. In: AAAI (2020)

\bibitem{ye2020feat}
Ye, H.J., Hu, H., Zhan, D.C., Sha, F.: Few-shot learning via embedding
  adaptation with set-to-set functions. In: CVPR (2020)

\bibitem{ye2020few}
Ye, H.J., Hu, H., Zhan, D.C., Sha, F.: Few-shot learning via embedding
  adaptation with set-to-set functions. In: CVPR (2020)

\bibitem{zhang2020deepemd}
Zhang, C., Cai, Y., Lin, G., Shen, C.: Deepemd: Few-shot image classification
  with differentiable earth mover's distance and structured classifiers. In:
  CVPR (2020)

\bibitem{zhang2019pyramid}
Zhang, C., Lin, G., Liu, F., Guo, J., Wu, Q., Yao, R.: Pyramid graph networks
  with connection attentions for region-based one-shot semantic segmentation.
  In: ICCV (2019)

\bibitem{zhang2019canet}
Zhang, C., Lin, G., Liu, F., Yao, R., Shen, C.: C{AN}et: Class-agnostic
  segmentation networks with iterative refinement and attentive few-shot
  learning. In: CVPR (2019)

\bibitem{zhang2020feature}
Zhang, D., Zhang, H., Tang, J., Wang, M., Hua, X., Sun, Q.: Feature pyramid
  transformer. In: ECCV (2020)

\bibitem{zhang2020sg}
Zhang, X., Wei, Y., Yang, Y., Huang, T.S.: S{G}-one: Similarity guidance
  network for one-shot semantic segmentation. IEEE Transactions on Cybernetics
  (2020)

\bibitem{zheng2021rethinking}
Zheng, S., Lu, J., Zhao, H., Zhu, X., Luo, Z., Wang, Y., Fu, Y., Feng, J.,
  Xiang, T., Torr, P.H., et~al.: Rethinking semantic segmentation from a
  sequence-to-sequence perspective with transformers. In: CVPR (2021)

\bibitem{zhengyu2021pareto}
Zhengyu, C., Jixie, G., Heshen, Z., Siteng, H., Donglin, W.: Pareto
  self-supervised training for few-shot learning. In: CVPR (2021)

\bibitem{zhiqiang2021partial}
Zhiqiang, S., Zechun, L., Jie, Q., Marios, S., Kwang-Ting, C.: Partial is
  better than all:revisiting fine-tuning strategy for few-shot learning. In:
  AAAI (2021)

\bibitem{zhong2017random}
Zhong, Z., Zheng, L., Kang, G., Li, S., Yang, Y.: Random erasing data
  augmentation. In: AAAI (2020)

\bibitem{zhu2020deformable}
Zhu, X., Su, W., Lu, L., Li, B., Wang, X., Dai, J.: Deformable detr: Deformable
  transformers for end-to-end object detection. In: ICLR (2020)

\end{thebibliography}
\end{document}